\documentclass[conference]{IEEEtran}
\IEEEoverridecommandlockouts

\usepackage{cite}
\usepackage{amsmath,amssymb,amsfonts}
\usepackage{algorithmic}
\usepackage{graphicx}
\usepackage{textcomp}
\usepackage{xcolor}
\usepackage{multirow}
\usepackage[colorlinks=true, linkcolor=green, citecolor=green, urlcolor=black]{hyperref}
\usepackage[final]{changes}

\def\BibTeX{{\rm B\kern-.05em{\sc i\kern-.025em b}\kern-.08em
    T\kern-.1667em\lower.7ex\hbox{E}\kern-.125emX}}

\begin{document}

\title{Chain-of-Thought Prompting for Speech Translation\\
}

\author{
    \IEEEauthorblockN{Ke Hu, Zhehuai Chen, Chao-Han Huck Yang, Piotr Żelasko, Oleksii Hrinchuk, Vitaly Lavrukhin, \\ Jagadeesh Balam, Boris Ginsburg}
    \IEEEauthorblockA{
        NVIDIA \\
        kevinhu@nvidia.com
    }
    \vspace{-3em}
}

\maketitle

\begin{abstract}
Large language models (LLMs) have demonstrated remarkable advancements in language understanding and generation. Building on the success of text-based LLMs, recent research has adapted these models to use speech embeddings for prompting, resulting in Speech-LLM models that exhibit strong performance in automatic speech recognition (ASR) and automatic speech translation (AST). In this work, we propose a novel approach to leverage ASR transcripts as prompts for AST in a Speech-LLM built on an encoder-decoder text LLM. The Speech-LLM model consists of a speech encoder and an encoder-decoder structure Megatron-T5. By first decoding speech to generate ASR transcripts and subsequently using these transcripts along with encoded speech for prompting, we guide the speech translation in a two-step process like chain-of-thought (CoT) prompting. Low-rank adaptation (LoRA) is used for the T5 LLM for model adaptation and shows superior performance to full model fine-tuning. Experimental results show that the proposed CoT prompting significantly improves AST performance, achieving an average increase of 2.4 BLEU points across 6 En$\to$X or X$\to$En AST tasks compared to speech prompting alone. Additionally, compared to a related CoT prediction method that predicts a concatenated sequence of ASR and AST transcripts, our method performs better by an average of 2 BLEU points. Our code is open-sourced on GitHub.\footnote{\url{https://github.com/kevinhu-nv/NeMo\_CoT}}
\vspace{-1em}
\end{abstract}

\section{Introduction}
\label{sec:intro}

Large language models (LLMs) have made rapid progress in the last couple of years \cite{chowdhery2023palm, brown2020language, team2023gemini, achiam2023gpt, touvron2023llama, bai2023qwen}. Built on billions of parameters and massive text data, LLMs have shown strong language understanding and generation abilities as well as emergent abilities such as in-context learning, instruction following, and multi-step reasoning.
Following the success of text LLMs, recent studies propose to adapt the text LLM to use speech embeddings for prompting \cite{wang2023slm, fathullah2024prompting, wu2023decoder, chen2024salm, lakomkin2024end, yang2024promptasr, kong2024audio, reid2024gemini, dubey2024llama}. By introducing speech as LLM prompting inputs, the Speech-LLM models show competitive performance in a number of speech tasks including automatic speech recognition (ASR) and automatic speech translation (AST).

Prompt design plays a critical role in leveraging the power of LLMs. In \cite{brown2020language}, it is shown that without fine tuning the model, one can use example contexts and completions as prompts, and ask the model to complete a new request. This form of in-context learning~\cite{min2022rethinking, wang2024can} ability highlights the importance of injecting guiding information into  prompts. Other approaches \cite{lester2021power, li2021prefix} append trainable embeddings to fixed ones to let the model learn the prompt via supervised training. Prompts with rich information may also help LLMs generate the correct response. For example, chain-of-thought (CoT) prompting \cite{wei2022chain} has shown to improve in a number of reasoning tasks such as math and reasoning. CoT prompting uses a multi-step prompting method to explore the LLM's generation ability and guide the model to the final answer.

Past work in various speech tasks also shows the benefits of a multi-step prediction. For example, deliberation models \mbox{\cite{le2022deliberation, hu2020deliberation, xia2017deliberation, yang2023generative}} first predict the first-pass hypothesis and then use that to assist a more sophisticated second-pass task. In the Speech-LLM framework, a joint audio and speech understanding model \cite{gong2023joint} uses Whisper \cite{radford2023robust} to generate spoken text and use that to prompt the LLaMA LLM \cite{touvron2023llama} for a range of audio tasks. Without using speech in prompting, \cite{hu2024gentranslate} develops a generic multi-task correction LLM takes outputs from various models and generates refined results. Recently, in \cite{huang2023speech}, a CoT prediction method for speech translation is proposed to predict a concatenated ASR and AST transcript sequence by prompting a decoder-only GPT. However, when using an encoder-decoder LLM such as T5, it is unclear what is the best place to inject the ASR hypothesis, i.e., decoder outputs (as in \cite{huang2023speech}) or T5 inputs, and is worth researching (note \cite{du2024cot}, \cite{chen2024llast} are developed concurrently with our work).

In this work, we investigate leveraging \textit{ASR transcripts in prompts} for speech translation based on an encoder-decoder text LLM. First, in an ASR task, we decode input speech to generate ASR transcripts, \textit{i.e.}, text in the source language. In principle, the ASR transcripts can be generated from any ASR system, and for fair comparison in this work, we implement the CoT prediction method \cite{huang2023speech} and take the ASR portion of the output as ASR transcripts. Then, for the proposed CoT prompting, we concatenate the AST textual prompt, previously generated ASR transcript, and speech encodings to a single sequence to prompt a Megatron-T5 LLM \cite{shoeybi2019megatron, raffel2020exploring} to generate text translations. We use a pretrained Canary encoder \cite{puvvada2024less} for speech encoding. Similar to \cite{brown2020language}, our model is trained by next token prediction loss. We always tune the speech encoder in training. For the LLM, given different performances in fine tuning techniques in previous works \cite{tang2023salmonn, chen2023lauragpt}, we compare full model fine tuning and LoRA \cite{hu2021lora} for the Megatron-T5 LLM, and results show that LoRA significantly improves our performance with minor parameter increase. Our experiments show that, in the encoder-decoder LLM framework, utilizing ASR transcripts in prompting significantly improves speech translation by an average of 2.4 BLEU score in 6 En$\to$X or X$\to$En AST tasks, compared to the baseline model without ASR transcripts in prompts. Compared to the CoT prediction method \cite{huang2023speech}, our method is around 2 BLEU score better on average. 

\section{Model Description}
\label{sec:model}

\begin{figure}[t]
    \centering
    \includegraphics[width=0.4\textwidth]{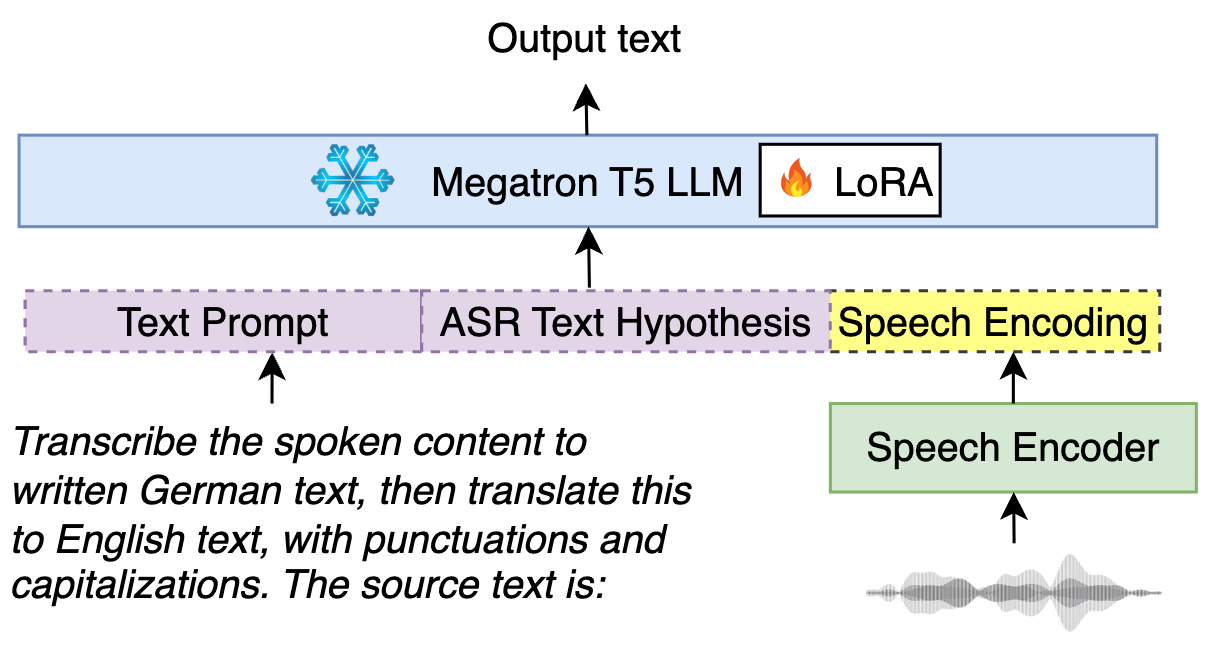}
    \caption{Diagram of the proposed chain-of-thought (CoT) prompting model. The fixed text prompt, ASR text hypotheses, and speech encodings are concatenated to a single sequence to prompt the Megatron-T5 for translation.}
    \label{fig:cot_diagram}
    \vspace{-2em}
\end{figure}

As shown in Fig. \ref{fig:cot_diagram}, our speech LLM consists of an audio encoder and a Megatron-T5 LLM. We use the audio encoder from Canary-1B \cite{puvvada2024less} pretrained for ASR and AST tasks. The Canary encoder is composed by 24 transformer layers and has around 650M parameters. The Megatron-T5 LLM is also pretrained and has an encoder-decoder structure and 1.2B parameters in total \cite{nvidia_riva_megatron_nmt_1b}. The input speech is first encoded by the Canary encoder and then prefixed by the text prompt in a single sequence as the input to the Megatron-T5 LLM. In particular, the text prompt contains two parts: 1) Fixed text prompt, and 2) The estimated ASR transcripts, i.e. ASR textual hypothesis.

In principal, the ASR hypotheses in our model can be obtained from any reasonable ASR systems. For a fair comparison with \cite{huang2023speech}, we first train a CoT prediction model as in \cite{huang2023speech}, which predicts a concatenated sequence of ASR and AST transcripts. We then take the ASR transcript portion of the output as the ASR text hypothesis in Fig. \ref{fig:cot_diagram}. For the CoT prediction model, we use a prompt such as ``\textit{Q: Transcribe the spoken content to written German text, then translate this to English text, with punctuations and capitalizations.\textbackslash nA:}" similar to \cite{huang2023speech}.

We train our CoT prompting model by injecting the ASR transcripts into the T5 prompts as shown in Fig. \ref{fig:cot_diagram}. The complete AST prompt looks like:
``\textit{Q: Transcribe the spoken content to written German text, then translate this to English text, with punctuations and capitalizations. The source text is:  \textbf{Verschandeln Sie die Stätte nicht durch Anbringen oder Einkratzen von Graffiti.}\textbackslash nA:}"
In this De$\to$En example, the predicted German ASR text is shown in bold. Note again that in our model the generation of the ASR hypotheses does not depend on \cite{huang2023speech}, and one can use any ASR model in practice. Without specific clarification, we fine tune the whole model in training. In experiments (Sect. \ref{sec:lora}), we have also applied LoRA \cite{hu2021lora} for more efficient model tuning.

The novelty of the proposed model is discussed by comparing to related systems below. Compared to the baseline model (same as Fig. \ref{fig:cot_diagram} but without ASR text hypothesis in prompts), our model performs translation by first predicting ASR hypothesis and then use that as input in addition to speech embedding for translation. The ASR prediction acts as the first step in a two-step translation process, similar to the CoT prompting in machine translation \cite{wei2022chain} where the model is given multiple steps of instructions to guide a task. The use of both speech and ASR predictions in the second-step prediction is similar to deliberation models \cite{hu2020deliberation, xia2017deliberation}, and here we further capitalize on the power of LLM by prompting. Compared to the CoT prediction method in \cite{huang2023speech}, our model uses an encoder-decoder structure T5 LLM instead of a decoder-only GPT in \cite{huang2023speech}. When using an encoder-decoder LLM architecture such as T5, it is unclear what is the best place to inject the ASR hypothesis (i.e., decoder outputs or T5 inputs) and is worth investigating. In addition, we present CoT AST results in multiple language pairs instead of English-Chinese translations in \cite{huang2023speech}. The effectiveness of injecting ASR hypotheses as T5 inputs is presented in Sect. \ref{sec:compare} by comparing to the CoT prediction method.

We train our CoT prompting model using the next token prediction loss \cite{brown2020language}. In decoding, the Megatron-T5 LLM takes the concatenation of text prompt, ASR transcripts and encoded audio as input, and then iteratively predict the next token by using the token at the current step as input.

\section{Experiment Details}
\label{sec:exp}

\subsection{Data}
\label{sec:data}


We use a subset of Canary AST training data \cite{puvvada2024less} to generate CoT training data for experiment efficiency (X$\to$En with 2,752, 1,795, and 1,397 hours for De, Fr, and Es, respectively, and En$\to$X with 1,640 hours for each language of De, Fr, and Es). We first generate the ASR hypothesis and then append the hypothesis to the fixed model prompt as T5 input. The combined textual prompt looks like: 

``\textit{Q: Transcribe the spoken content to written $\{source\_lang\}$ text, then translate this to $\{target\_lang\}$ text, with punctuations and capitalizations. The source text is: \{ASR\_transcript\}\textbackslash nA:}"

Here, $\{source\_lang\}$ and $\{target\_lang\}$ are source and target language names, respectively, and $\{ASR\_transcript\}$ represents the estimated ASR text hypothesis for the utterance. The punctuation and capitalization prompts are applied depending on the target text. We have generated ASR hypotheses as described in Sect. \ref{sec:model} for 3 source languages in X$\to$En, and for English in En$\to$X translations. Our AST target labels are generated synthetically by NVIDIA Megatron NMT models \cite{nvidia_megatronnmt_any_en_500m, nvidia_megatronnmt_en_any_500m}. In inference, we  evaluate AST model performance using the FLEURS \cite{conneau2023fleurs} test sets. We first generate ASR hypothesis in the same way as in training and then inject them in AST prompt for inference.

\subsection{Modeling Details}
We implement the model with PyTorch using NeMo Toolkit \cite{kuchaiev2019nemo}, and the model is trained on 32 A100 (80G) GPUs with a batch duration of 180 sec per GPU. The speech encoder is initialized from the 20-layer Canary-1B encoder \cite{puvvada2024less}, and the LLM is initialized from the 1.2B Megatron-T5 NMT model \cite{nvidia_riva_megatron_nmt_1b}. The T5 LLM consists of a 12-layer encoder and a 24-layer decoder, with 20 attention heads and a hidden dimension of 1280, and the feedforward layer has a dimension of 5120. Relative positional embedding is used. We use a 64k SentencePiece tokenizer for all languages. RMSNorm \cite{zhang2019root} is used for normalization for all layers of the T5 model. We use fused Adam, and an inverse Square Root Annealing learning rate (LR) schedule for optimization. The LR schedule starts with an initial learning rate of 4e-4. Gradient clipping is applied at a threshold of 1.0 to stabilize training. Warmup steps are configured to 0.8\% of 3.6M maximum steps. 

The proposed CoT model has a total parameter size of around 1.8B. In experiments, we have tried both full model fine tuning and LoRA \cite{hu2021lora} for LLM adaptation. For the Speech-LLM model optimization, we use distributed fused Adam optimizer and the cosine annealing LR scheduler with a learning rate of 1e-4 and no weight decay. Gradient clipping of 1.0 is applied.

\section{Results}
\label{sec:results}

\subsection{Chain-of-Thought (CoT) prompting}
\label{sec:cot_prompt}

\begin{table}[t]
\caption{The effect of CoT prompting using estimated ASR hypotheses.}
\centering
\begin{tabular}{|c c c|}
\hline
Lang. Pair & SALM-T5 Baseline & CoT Prompt (\texttt{E1}) \\ \hline \hline
De$\to$En & 36.6 & \textbf{37.6} \\ \hline
Fr$\to$En & 33.9 & \textbf{35.6} \\ \hline
Es$\to$En & 24.6 & \textbf{25.6} \\ \hline
En$\to$De & 29.8 & \textbf{31.9} \\ \hline
En$\to$Fr & 40.1 & \textbf{41.6} \\ \hline
En$\to$Es & 21.8 & \textbf{23.1} \\ \hline
Avg. & 31.1 & \textbf{32.6} \\ \hline
\end{tabular}
\label{table:cot_prompt}
\vspace{-1em}
\end{table}

\begin{table}[t]
\centering
\caption{CoT prompting using ASR hypotheses or ground truth labels.}
\begin{tabular}{|c c c|}
\hline
Lang. Pair & Prompt w/ hyp (\texttt{E1}) & Prompt w/ GT \\ \hline \hline
De$\to$En & 37.6 & \textbf{40.2} \\ \hline
Fr$\to$En & 35.6 & \textbf{39.5} \\ \hline
Es$\to$En & 25.6 & \textbf{28.1} \\ \hline
En$\to$De & 31.9 & \textbf{34.3} \\ \hline
En$\to$Fr & 41.6 & \textbf{44.6} \\ \hline
En$\to$Es & 23.1 & \textbf{25.1} \\ \hline
Avg. & 32.6 & \textbf{35.3} \\ \hline
\end{tabular}
\label{table:cot_prompt_gt}
\vspace{-2em}
\end{table}

We first evaluate the performance of the CoT prompting model, i.e., injecting the ASR hypotheses to the LLM prompt. As shown in Table \ref{table:cot_prompt}, the CoT prompt (\texttt{E1}) benefits translation for all language pairs and achieved an average improvement of 1.5 BLEU score compared to the baseline (from 31.1 $\to$ 32.6). The baseline model is trained in the same way as the CoT prompt model, except removing ASR hypotheses and using the prompt such as: ``\textit{Translate the spoken $\{source\_lang\}$ content to written $\{target\_lang\}$ text, with punctuations and capitalizations.}" We call the baseline model SALM-T5 since the prompt concatenation follows the same way as the SALM \cite{chen2024salm}. Our improvement is for both En$\to$X and X$\to$En translations. In this experiment, both training and inference have used estimated ASR hypotheses. The ASR hypotheses are generated and then appended to the AST prompt as described in Sect. \ref{sec:data}.

\begin{table}[t]
\centering
\caption{CoT prediction by training using ground truth ASR labels or estimated ASR hypotheses.}
\begin{tabular}{|c c c|}
\hline
Lang. Pair & Train w/ ASR GT (B2) & Train w/ ASR hyp \\ \hline \hline
De$\to$En & 34.2 & \textbf{35.2} \\\hline
Fr$\to$En & \textbf{35.1} & 34.1 \\\hline
Es$\to$En & \textbf{25.5} & 25.1 \\\hline
En$\to$De & \textbf{31.0} & 30.5 \\\hline
En$\to$Fr & 40.4 & \textbf{40.8} \\\hline
En$\to$Es & \textbf{22.8} & 22.4 \\\hline
Avg. & \textbf{31.5} & 31.4 \\ \hline
\end{tabular}
\label{table:cot_predict}
\vspace{-1.5em}
\end{table}

\begin{table}[t]
\centering
\caption{Adding LoRA to the CoT prompting model.}
\begin{tabular}{|c c c|}
\hline
Lang. Pair & E1 & E1 + LoRA (\texttt{E2}) \\ \hline \hline
De$\to$En & 37.6 & \textbf{38.3} \\ \hline
Fr$\to$En & 35.6 & \textbf{36.6} \\ \hline
Es$\to$En & 25.6 & \textbf{26.7} \\ \hline
En$\to$De & 31.9 & \textbf{32.6} \\ \hline
En$\to$Fr & 41.6 & \textbf{43.4} \\ \hline
En$\to$Es & 23.1 & \textbf{23.4} \\ \hline
Avg. & 32.6 & \textbf{33.5} \\ \hline
\end{tabular}
\label{table:lora}
\vspace{-2em}
\end{table}

\subsection{Prompt with ground truth ASR transcripts}

To measure the impact of estimated ASR hypothesis quality on CoT prompting, we have tried using ground truth ASR labels in prompts. This is to evaluate whether there is potential improvement by using a better ASR model. As indicated in Table \ref{table:cot_prompt_gt}, we have obtained an average of 2.7 BLEU score improvement (32.6 $\to$ 35.3) for all languages by using the ground truth ASR labels, compared to \texttt{E1} which uses estimated ASR hypotheses. It means that better quality ASR prediction does benefit translation. On the other hand, it will be interesting to see how the model performs with lower quality ASR transcripts as inputs, i.e., maybe ones generated from a lightweight and efficient model as the first pass.

\begin{table*}[t]
\centering
\caption{Comparison of baseline SALM-T5 model, CoT prediction \cite{huang2023speech}, SeamlessM4T \cite{barrault2023seamlessm4t}, and the proposed CoT prompting with LoRA.}
\begin{tabular}{|c l c c c c c c c|}
\hline
\multirow{2}{*}{ID} & \multirow{2}{*}{Model} & \multicolumn{6}{c}{BLEU} & \multirow{2}{*}{Avg. BLEU} \\ \cline{3-8}
 &  & De$\to$En & Fr$\to$En & Es$\to$En & En$\to$De & En$\to$Fr & En$\to$Es & \\ \hline \hline
B1 & SALM-T5 Baseline & 36.6 & 33.9 & 24.6 & 29.8 & 40.1 & 21.8 & 31.1 \\ \hline
B2 & CoT Prediction & 34.2 & 35.1 & 25.5 & 31.0 & 40.4 & 22.8 & 31.5 \\ \hline
B3 & SeamlessM4T-medium \cite{barrault2023seamlessm4t} & 33.4 & 31.0 & 21.7 & 28.3 & 37.4 & 21.1 & 28.8 \\ \hline
B4 & SeamlessM4T-large-v2 \cite{barrault2023seamlessm4t} & 37.1 & 30.9 & 25.4 & \textbf{33.2} & 43.1 & \textbf{23.7} & 32.2 \\ \hline
E1 & CoT Prompting & 37.6 & 35.6 & 25.6 & 31.9 & 41.6 & 23.1 & 32.6 \\ \hline
E2 & E1 + LoRA & \textbf{38.3} & \textbf{36.6} & \textbf{26.7} & 32.6 & \textbf{43.4} & 23.4 & \textbf{33.5}\\ \hline
\end{tabular}%
\label{table:comparison}
\vspace{-1.5em}
\end{table*}

\begin{table*}[t]
\centering
\caption{Comparison of a cascade system and the CoT+LoRA prompting model.}
\begin{tabular}{|c l c c c c c c c|}
\hline
\multirow{2}{*}{ID} & \multirow{2}{*}{Model} & \multicolumn{6}{c}{BLEU} & \multirow{2}{*}{Avg. BLEU} \\ \cline{3-8}
 &  & De$\to$En & Fr$\to$En & Es$\to$En & En$\to$De & En$\to$Fr & En$\to$Es & \\ \hline \hline
B5 & Cascade NMT & \textbf{38.4} & \textbf{36.7} & 26.6 & 30.1 & 42.2 & 22.9 & 32.8 \\ \hline
E2 & CoT Prompting + LoRA & 38.3 & 36.6 & \textbf{26.7} & \textbf{32.6} & \textbf{43.4} & \textbf{23.4} & \textbf{33.5} \\ \hline
\end{tabular}%
\label{table:comp_cascade}
\vspace{-2em}
\end{table*}

\subsection{CoT prediction}
\label{sec:cot_pred}

Since we use an encoder-decoder T5 instead of the decoder-only GPT, we investigate what is the best place to inject ASR hypotheses. In addition to injecting in the T5 encoder (Sect. \ref{sec:cot_prompt}), we have also tried predicting ASR hypotheses first and then followed by AST output (i.e., \cite{huang2023speech}). Similar to \cite{huang2023speech}, we use the following prompt: ``\textit{Q: Transcribe the spoken content to written \{source\_lang\} text, then translate this to \{target\_lang\} text, with punctuations and capitalizations.}" We have experimented two setups in this experiment: 1) ASR ground truth text is used in the target sequence for prediction (same as \cite{huang2023speech}), or 2) estimated ASR hypotheses are used target sequence to create a more matched condition as in inference (second column in Table \ref{table:cot_predict}). When using ground truth ASR hypotheses, the model predicts the concatenated ASR and AST hypotheses in a single sequence. We use a special separator token to concatenate the two labels, and the loss is calculated for the whole sequence. When using the predicted ASR hypotheses, we mask the loss over the ASR part and only use the loss from the AST prediction. In either scenario (in Table \ref{table:cot_predict}), we have not observed better performance of the CoT prediction method compared to the CoT prompting method. 

\subsection{Low-rank adaptation (LoRA) performance}
\label{sec:lora}

We have experimented adding LoRA to the Megatron-T5 model and achieved significant improvements across all languages (Table \ref{table:lora}). Since the Megatron T5 has an encoder-decoder structure, LoRA adapters have been added to both the encoder and decoder. We use an adapter dimension of 128 for all the 12 self-attention layers in the encoder. For the decoder, we have added LoRA adapters to both self-attention and cross attention layers for every decoder layer. For self-attention layers, we use the same adapter dimension of 128, and for cross-attention, we use an adapter dimension of 32 for queries and 64 for keys and values. In total, this adds 8M and 47M parameters to the encoder and decoder, respectively. As shown in Table \ref{table:lora}, the LoRA adapter significantly improves the model performance by improving the BLEU score from 32.6 $\to$ 33.5. The improvement ranges from 0.3 to 1.8 BLEU. LoRA seems to maintain the translation ability of the original text LLM, which may be beneficial in our scenario with text ASR transcripts in the prompt.

\subsection{Comparisons}
\label{sec:compare}

In Table \ref{table:comparison}, we compare our CoT prompting models (\texttt{E1} and \texttt{E2}) to several  models including the SALM-T5 baseline (\texttt{B1}), a CoT prediction model (\texttt{B2}), and SeamlessM4T models (\texttt{B3} and \texttt{B4}). Note that we have implemented the CoT prediction \cite{huang2023speech} based on the encoder-decoder T5 LLM instead of the decoder-only GPT in \cite{huang2023speech}. For SeamlessM4T \cite{barrault2023seamlessm4t}, we have used the official checkpoints to rerun the model. We evaluate the models by using the FLEURS dataset \cite{conneau2023fleurs} for translation of 6 language pairs (3 X$\to$En and 3 En$\to$X).

We first compare the SALM-T5 baseline (\texttt{B1}) and the CoT prompting (\texttt{E1}), and the performance difference is only due to adding ASR hypotheses to the T5 prompts. We see in Table \ref{table:comparison} that the latter performs 1.5 BLEU score better on average. Then, by adding LoRA adapters, our CoT prompting model performs an additional 0.9 BLEU better, achieving an average of 2.4 BLEU better compared to the baseline (\texttt{B1}). We can see the improvement is uniform for all language pairs. Secondly, we compare to the CoT prediction method in \cite{huang2023speech}. Compared to the baseline \texttt{B1}, the results show that CoT prediction \cite{huang2023speech} improves the baseline by around 0.4 BLEU. It is effective for most language pairs but worse in De$\to$En translations. Comparing CoT prediction (\texttt{B2}) to the proposed CoT prompting method (\texttt{E1}), the injection of ASR hypotheses in prompts improves the BLEU by 0.9 on average (31.5 $\to$ 32.6). This improvement shows that injecting the ASR transcripts as T5 inputs is more effective than decoder outputs. Note that we have used the same ASR hypotheses in both models. Lastly, to put model performance into perspective, we compare our best performing model (\texttt{E2}, with LoRA adapters) to the SeamlessM4T medium and large models \cite{barrault2023seamlessm4t}. Our LoRA model performs 4.7 and 1.3 BLEU better than the medium and large SeamlessM4T models, respectively. The SeamlessM4T medium and large models have sizes of 1.2B and 2.3B, respectively, while our model has a total size of 1.8B. However, we also note that the proposed model only performs speech translation but SeamlessM4T is capable of a range of speech and text tasks.

To measure the effect of using both speech and ASR text transcription for prompting, we further compare our LoRA model (\texttt{E2}) to a cascade system. In the cascade system, the ASR hypotheses are first generated and then a machine translation model is used for text translation. We have used the same ASR hypotheses as those for CoT prompting in \texttt{E2} for fair comparison. For machine translation, we have used two separate Megatron 500M models for En-to-any \cite{nvidia_megatronnmt_en_any_500m} and any-to-En \cite{nvidia_megatronnmt_any_en_500m} translations, totaling 1B model parameters. We have tried using a 1B any-to-any MT model for translation but achieved worse results. In Table \ref{table:comp_cascade}, we can see that our system performs similarly in 3 X$\to$En language pairs, and 0.5-2.5 BLEU score better for En$\to$X translations. The better performance of our model in translating English to other languages is probably due to initializing from the Canary encoder.

\section{Conclusions}
\label{sec:conclude}

We have investigated chain-of-thought (CoT) prompting for a SpeechLLM built on an encoder-decoder architecture. 
Our best performing model achieved an average BLEU score improvement of 2.4 points compared to the SALM-T5 baseline. Compared to a CoT prediction model similar to \cite{huang2023speech}, our method performs 2 BLEU score better across all language pairs. The effectiveness of using both speech and ASR text in prompting is demonstrated by a gain of up to 2.5 BLEU over a traditional cascade system.

\bibliographystyle{IEEEtran}
\footnotesize
\bibliography{ref}

\end{document}